# GeoFlow-SLAM: A Robust Tightly-Coupled RGBD-Inertial and Legged Odometry Fusion SLAM for Dynamic Legged Robotics

Tingyang Xiao[1], Xiaolin Zhou[1], Liu Liu[1], Wei Sui[2], Wei Feng[1], Jiaxiong Qiu[1], Xinjie Wang[1], and Zhizhong Su[1]

***Abstract*—This paper presents GeoFlow-SLAM, a robust and effective Tightly-Coupled RGBD-Inertial and Legged Odometry Fusion SLAM for legged robotics undergoing aggressive and high-frequency motions. By integrating geometric consistency, legged odometry constraints, and dual-stream optical flow (GeoFlow), our method addresses three critical challenges: feature matching and pose initialization failures during fast locomotion and visual feature scarcity in texture-less scenes. Specifically, in rapid motion scenarios， feature matching is notably enhanced by leveraging dual-stream optical flow, which combines prior map points and poses. Additionally, we propose a robust pose initialization method for fast locomotion and IMU error in legged robots, integrating IMU/Legged odometry, inter-frame Perspective-n-Point (PnP), and Generalized Iterative Closest Point (GICP). Furthermore, a novel optimization framework that tightly couples depth-to-map and GICP geometric constraints is first introduced to improve the robustness and accuracy in long-duration, visually texture-less environments. The proposed algorithms achieve state-of-the-art (SOTA) on collected legged robots and open-source datasets. To further promote research and development, the open-source datasets and code will be made publicly available at https://github.com/HorizonRobotics/GeoFlowSlam.**

## I. Introduction

The RGB-D camera can capture RGB information while simultaneously providing depth information for all areas within its field of view, regardless of whether these areas have textures or not [1]. This capability makes the RGBD camera a popular sensor in robotics, particularly in applications like Visual Simultaneous Localization and Mapping (SLAM).

For legged robots, fast motion jitter often occurs due to variations in speed and posture associated with each movement cycle, resulting in image blur and IMU distortion. Besides, these robots typically operate in texture-less indoor environments. These present new challenges for legged robot localization. However, RGB/RGBD-based SLAM [2], [3], [4], [5],[6], [7] cannot effectively address these issues. Specifically, fast motion jitter induces image blur, which subsequently complicates feature matching. When IMU is unused, or when IMU distortion occurs due to the unique motion characteristics of legged robots [8], [9], the robot's pose usually cannot be accurately initialized in fast locomotion. Furthermore, in long-duration operations within texture-less environments, depth information is primarily used to estimate the depth of feature points, without supplementary constraints that could enhance the overall accuracy and robustness of the system. To address the above problems, we propose GeoFlow-SLAM, with the following key contributions:

1) A dual-stream optical flow tracking incorporating map points and pose information is proposed to significantly enhance feature matching, particularly in fast locomotion scenarios.
2) A pose initialization method based on IMU/Legged Odometry, Perspective-n-Point (PnP), and Generalized Iterative Closest Point (GICP) is developed to address pose initialization failures during fast locomotion.
3) A novel optimization framework is first proposed that tightly integrates depth-to-map and GICP geometric consistency, improving robustness and accuracy, especially in texture-less environments.
4) Our algorithm achieves state-of-the-art (SOTA) on open-source datasets and datasets we collected using legged robotics. The open-source datasets on legged robotics and code will be released to facilitate further research and development.

## II. Related Work

Kinect-Fusion [10] is a real-time 3D reconstruction system based on RGB-D cameras， which generates point clouds from depth images, and estimates the camera pose through Iterative Closest Point (ICP). Kintinuous [11] builds upon Kinect-Fusion by adding loop closure detection and using GPU acceleration for pose estimation. Elastic-Fusion [12] leverages the color consistency of RGB to estimate the camera pose, in addition to employing depth-based ICP for pose estimation. However, they focus on 3D reconstruction rather than pose estimation.

ORBSLAM [5], VINS-RGBD [7], and RTAB-Map [13] are feature points-based RGB-D SLAM systems, where the depth information is mainly used to recover the depth of feature points, without providing additional observational constraints. In ORB-SLAM-based RGB-D SLAM systems, feature tracking predominantly relies on the depth and pose prior information of feature points. Conversely, in VINS-based systems, optical flow tracking does not utilize depth and pose prior information to assist tracking. In scenarios characterized by rapid motion, feature points are highly susceptible to tracking failure.

DS-SLAM [6], RSO-SLAM [14], and ViQu-SLAM [8] utilize depth information to assist in identifying and removing dynamic feature points in the environment, thereby enhancing localization accuracy. Based on the VINS and ORBSLAM frameworks, some RGB-D SLAM systems for structured environments have emerged, such as PLD-VINS [15], PLP-

[1]Horizon Robotics, Beijing, China
[2]D-Robotics, Beijing, China

SLAM [16], S-VIO [17], STL-SLAM [18], and VID-SLAM [19]. Although the extraction of structural features enhances localization accuracy and robustness to some extent, these methods are highly dependent on the environment. In unstructured environments, their robustness is significantly compromised. Moreover, the additional structural feature extraction is computationally expensive and time-consuming.

Nerf-based and Gaussian-based RGB-D SLAM systems such as NICE-SLAM [20], and SplaTAM [21], focus on dense reconstruction of 3D scene geometry and texture, which generally do not achieve real-time performance. In the absence of visual observations, IMU (Inertial Measurement Unit), which does not rely on environmental features, can provide continuous pose output and pre-integration constraints and is widely used in various RGB-D SLAM algorithms (e.g., VINS-RGBD, ORBSLAM3, VID-SLAM, DDIO-Mapping [22], S-VIO, etc.). However, due to the presence of IMU zero-bias noise and the lack of depth constraints, pose divergence is still inevitable in environments where visual feature constraints are missing for a long time, affecting the accuracy and robustness of the system. The R3LIVE [23] and FASTLIVO [24], [25] series of algorithms tightly couple IMU, Visual, and LiDAR observations, but they are not suitable for RGB-D cameras.

Existing RGB-D-Inertial SLAM systems often fail to effectively track features and fully utilize depth information, resulting in poor robustness in unstructured environments. Moreover, variations in speed and morphology during motion can disrupt the IMU in legged robots, rendering traditional visual-motion fusion methods unsuitable [8].

In the field of legged robot SLAM, MIT Cheetah2 [26] and ANYmal [27] use a combination of stereo, RGB-D cameras, or LiDAR to position and build robot-centric maps. Based on LIOSAM [28], Leg-KILO [9] presents a novel leg odometry using an on-manifold error-state Kalman filter (ESKF), which tightly couples leg kinematics and IMU. ViQu-SLAM addresses the challenges posed by moving objects in dynamic environments, enabling the integration of leg odometry (LO) from a legged robot and visual-inertial odometry (VIO). However, the above methods for legged robots rely on costly and bulky LiDAR sensors, and they fail to adequately address real-world challenges. Specifically, these systems struggle in texture-less environments and during fast locomotion, leading to feature tracking failures and decreased positioning accuracy.

## III. System Overview

The proposed GeoFlow-SLAM system is composed of three components: Feature Extraction and Tracking, IMU/Legged/PnP/GICP Odometry, and Visual-Depth-IMU Integration Mapping, as depicted in Figure 1.

### 1) *Feature Extraction and Tracking*

Depth plane features are initially extracted from depth point clouds and employed in the GICP algorithm to obtain an initial pose estimate. Subsequently, Oriented FAST and Rotated BRIEF (ORB) features are extracted from image data. Dual-Stream Optical Flow Tracking and the Perspective-n-Point (PnP) algorithm are then employed for feature tracking. The IMU raw measurements and legged encoder are pre-integrated, providing the foundation for IMU/Legged Odometry.

### 2) *IMU/Legged/PnP/GICP Odometry*

When both legged and IMU odometry are available, they are the primary used for pose prediction. In their absence, solving the PnP problem across successive frames determines the robot's initial pose. This initial pose from IMU/Legged or PnP is further refined using the GICP algorithm. After pose initialization and feature tracking, the depth-to-map registration and the reprojection of visual map points are employed to generate the observation residuals.

### 3) *Visual-Depth-IMU Integration Mapping*

This module comprises single frame optimization, local keyframe optimization, and global keyframe optimization. In the single frame pose optimization and local keyframe bundle adjustment (BA) module, visual reprojection residuals, depth-to-map residuals, and IMU/Legged/GICP pose constraints are tightly coupled within the factor graph framework. Different from single frame pose optimization, which focuses solely on pose estimation of current adjacent frames, local BA estimates the pose of local keyframes and the corresponding visual map points. When a loop closure is detected, the overall keyframe poses and the visual point cloud map are further optimized in the global optimization module.

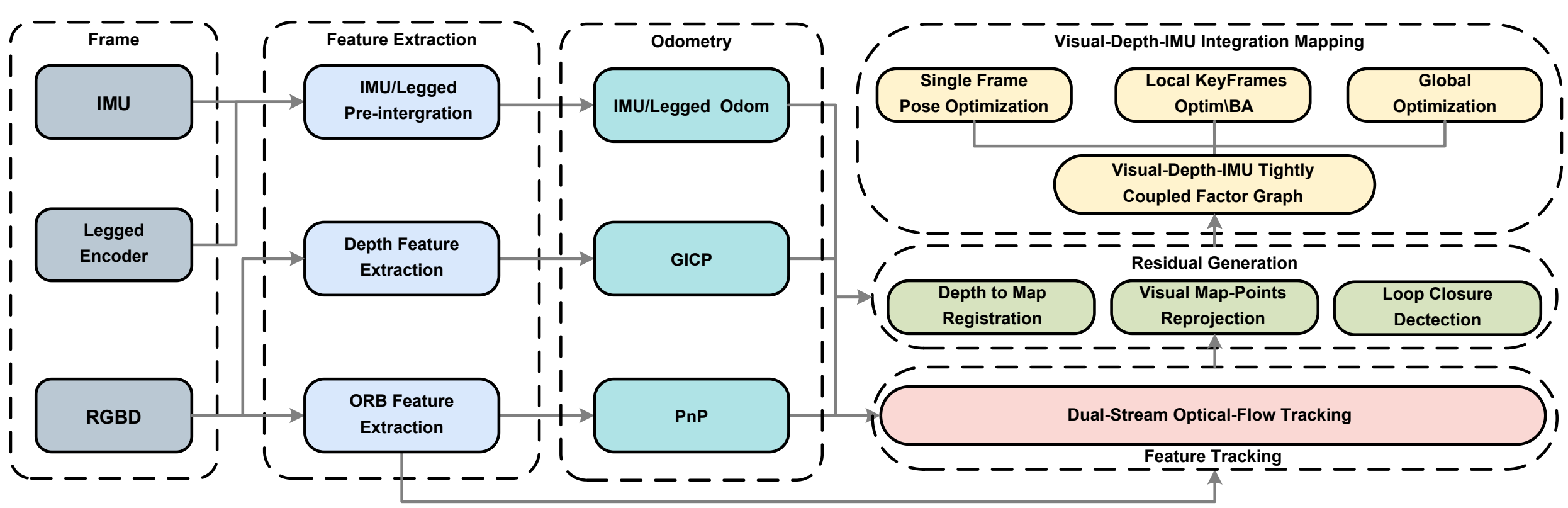

Figure 1. Overview of Our GeoFlow-SLAM Pipeline. Initially, utilizing data from RGBD, IMU, and Legged Encoders, we perform the extraction of depth plane features, and ORB features, and carry out IMU/legged pre-integration. The initial pose is determined using GICP, PnP, and IMU/Legged Odometry. Building upon this preliminary pose, Dual-Stream Optical Flow is subsequently utilized to track ORB features. Subsequently, single-frame and local keyframes are optimized within a tightly coupled Vision-Depth-IMU factor graph, enabling accurate map construction and pose estimation.

## IV. Method

### A. Dual-Stream Optical Flow Tracking Aided by Prior Map Points and Pose

Traditional optical flow generally operates under the assumption that motion is small and continuous. However, when in rapid motion, optical flow estimates based on the brightness constancy assumption can become inaccurate or fail to converge. Moreover, computing optical flow directly on high-resolution images leads to slow algorithm convergence and significant computational resource consumption. To address these challenges, we propose a dual-stream optical flow tracking method, as illustrated in Figure 2. , which utilizes prior map points (corresponding map points of ORB feature) and poses to enhance feature tracking robustness during rapid motion. This approach effectively mitigates the common issue of feature tracking failure, which often occurs when pixels undergo large displacements due to fast camera motion.

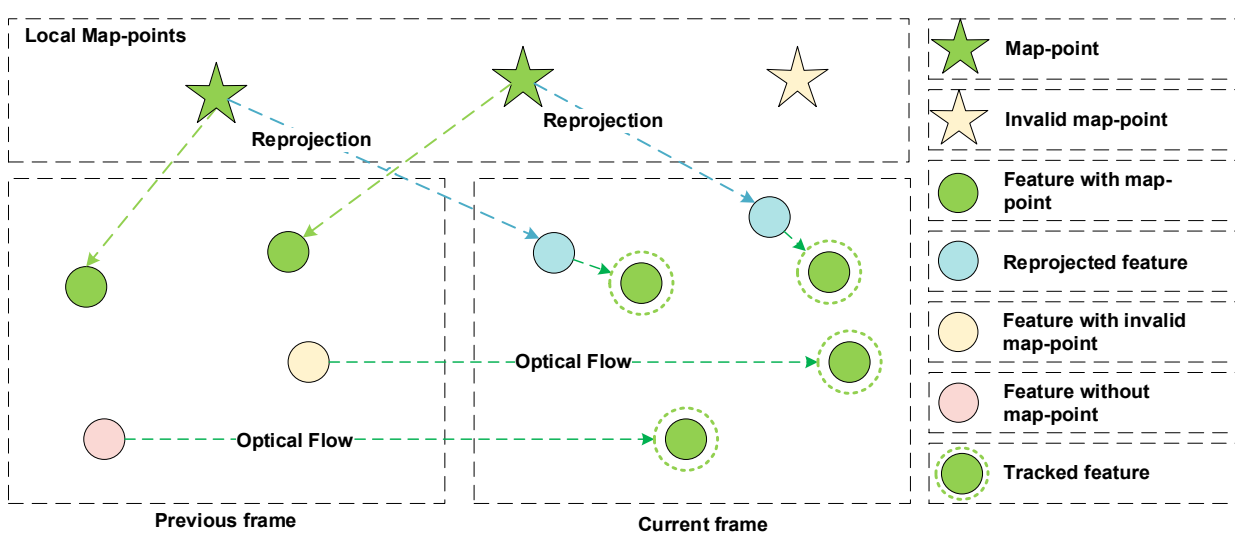

Figure 2. The prior map points and pose-aided dual-stream optical flow pipeline. Initially, the feature points corresponding to prior map points are subjected to reprojection, generating the reprojected points (depicted as blue points). Subsequently, optical flow is conducted on these reprojected points based on the current frame. Next, optical flow tracking is further applied to the feature points in the previous frame that lack qualified prior map points (represented as yellow and pink points).

#### 1) 3D-to-2D Optical Flow Tracking

In the first stream, we project the corresponding map points $\boldsymbol{P}(X,Y,Z)$ of feature points $p(x,y)$ into the current frame $p'(x,y)$ using the initial pose provided by IMU/Legged Odometry or PnP, GICP in odometry module.

$$p'(x,y) = F(\boldsymbol{T}_w^{C_{k-1}}\boldsymbol{P}) \tag{1}$$

$$F(\boldsymbol{P}) = \begin{pmatrix} f_X \frac{X}{Z} + c_x \\ f_Y \frac{Y}{Z} + c_y \end{pmatrix} \tag{2}$$

where $F(\boldsymbol{P})$ represents the intrinsic projection. Then, 3D-to-2D optical flow tracking is conducted on the projected pixels $p'(x,y)$ based on the current frame. $\boldsymbol{T}_w^{C_{k-1}}$ denotes the camera pose in time $k-1$. By leveraging precise priors of motion and map points for feature points, we achieve enhanced matching accuracy during rapid motion. Moreover, this method inherently filters out unstable features, such as dynamically inconsistent features.

#### 2) 2D-to-2D Optical Flow Tracking

While the first stage offers advantages, ORB features often lack valid corresponding map points, especially in cases where triangulation fails. Additionally, an inaccurate initial pose can significantly reduce the success rate of 3D-to-2D feature tracking. To address these challenges, a secondary stream is introduced, employing 2D-to-2D optical flow tracking on the original feature points. This additional step ensures continuous tracking, even in the absence of depth information. To further improve robustness, a fundamental matrix is computed to identify and eliminate outliers, accompanied by an inlier masking mechanism that prevents the unnecessary inclusion of tracking points within the same region. This combined approach effectively mitigates the impact of outliers, ensuring a more reliable tracking process.

As shown in Figure 3. , while ORB-SLAM3 fails to track effective feature points, our dual-stream optical flow tracking algorithm successfully tracks features such as those on the ground and the door latch, even under challenging conditions like motion blur in fast locomotion or in texture-less environments, such as large white walls.

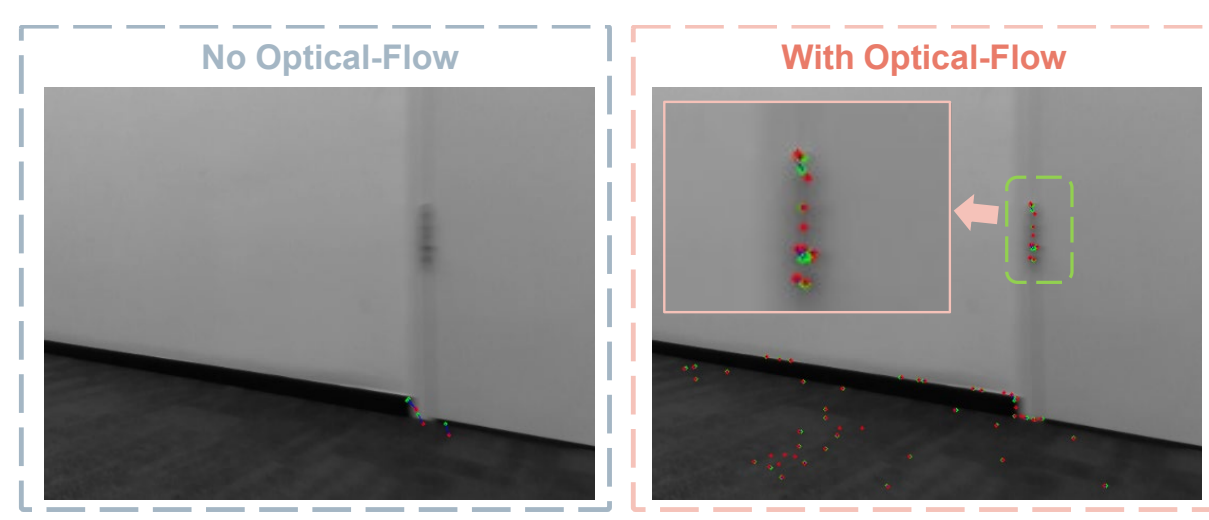

Figure 3. Feature tracking result based on dual-stream optical flow. The left image is the ORBSLAM3 result and the right is ours. The green points denote feature points, while the red points indicate the reprojected points onto the image.

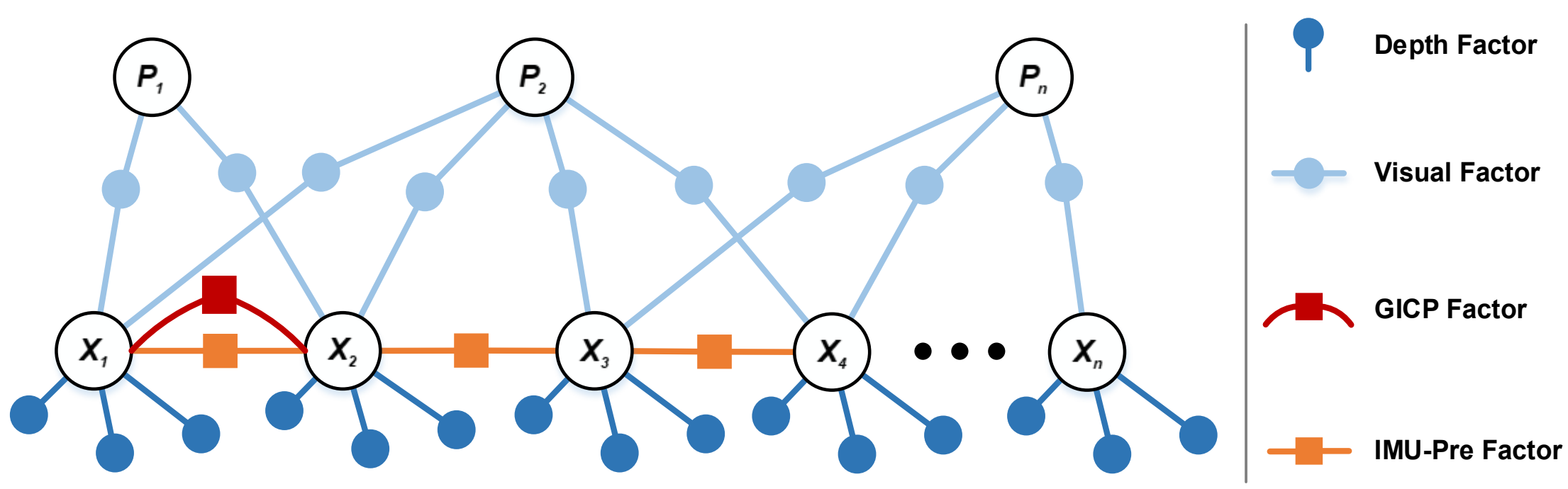

Figure 4. Factor graph framework of GeoFlow-SLAM in local keyframe optimization. The visual map point $P_1 - P_n$ and pose $x_1 - x_n$ in a window are jointly estimated by integrating visual reprojection residual, depth to map residual, GICP residual, and IMU pre-integration residual. To enhance computational efficiency, we only integrate depth-to-map constraints when valid visual observations are insufficient.

### B. Legged, PnP, and GICP Odometry

In a texture-less environment, the inability to obtain an accurate and robust initial pose compromises the effectiveness of optical flow tracking and the accuracy of visual reprojection and depth-to-map data association, ultimately leading to system divergence. Moreover, for legged robots, variations in speed and morphological changes during each motion mode cycle can disrupt the robot's IMU [8], [9], resulting in erroneous initial values during integration and ultimately causing system divergence. To obtain an accurate and robust initial pose, we have developed an initial pose estimation method that integrates the Legged Odom, PnP, and GICP.

First, if Legged Odom data is available, it is prioritized for pose prediction to avoid incorrect initial pose predictions caused by IMU errors in legged robots. When the IMU is available, given its high accuracy over short durations and relatively simple computation, we prioritize using the IMU for initial pose prediction. In scenarios where the IMU is unused, we initially compute the Essential Matrix based on successfully tracked feature points to recover the frame-to-frame pose. However, this method does not provide scale information, prompting us to further conduct PnP based on depth to achieve a more accurate initial pose. If PnP fails, the initial pose is further refined by GICP. If all the aforementioned methods fail, the initial pose is estimated based on the constant velocity model assumption.

For enhanced computational efficiency, we utilize small_gicp [29] for GICP computation. Simultaneously, GICP-based relative pose constraints between consecutive frames are incorporated within the factor graph optimization framework to further improve accuracy.

### C. Tightly Coupled Visual, Depth, and IMU

In texture-less environments, robust and accurate positioning is often challenging for most RGBD SLAM systems that heavily rely on RGB information. Depth information, however, can provide additional structural details that are beneficial to systems. Therefore, it is crucial to enhance the performance of RGBD SLAM to integrate depth observation further. As shown in Figure 4. , we construct the following factors graph to optimize the poses of keyframes and visual feature points within a local window. This graph comprises visual reprojection residual factors, depth-to-map point-to-plane residual factors, GICP relative pose residual factors, and IMU pre-integration factors. The depth plane feature points of keyframes are stitched to form a local map. For each plane feature point in the current keyframe, plane fitting is conducted on the k-nearest point clouds after projecting the point onto the local map, resulting in point-to-plane residuals. The inter-frame pose is subsequently constrained using the GICP. Specifically, if the IMU pre-integration significantly deviates from the legged odometry or the constant-velocity assumption, indicating IMU integration divergence, we adaptively introduce legged odometry constraints to improve the pose estimation.

After the optimization, we utilize the optimized camera poses to reconstruct a more accurate local depth map. To ensure the real-time performance of the system, we only add depth-to-map constraints to the frame node when valid visual observations are insufficient. This approach ensures both the accuracy and real-time performance of the system.

For single-frame optimization, we also follow ORB-SLAM by constructing constraints between adjacent frames, including IMU pre-integration, GICP relative pose, visual reprojection, and depth-to-map constraints.

$$r_1 = \min_X \sum_{(i,j)\in F} \rho\left(\|r_v(p^{c_i}, X)\|^2_{\Sigma_v}\right) \tag{3}$$

$$r_2 = \min_X \sum_{i\in F} \rho\left(\left\|r_{imu}\left(z_{b_i b_{i+1}}, X\right)\right\|^2_{\Sigma_{b_i b_i+1}}\right) \tag{4}$$

$$r_3 = \min_X \sum_{(i,j)\in F} \rho\left(\left\|r_{icp}\left(I^{c_i}_{c_j}, X\right)\right\|^2_{\Sigma_{icp}}\right) \tag{5}$$

$$r_4 = \min_X \sum_{(i,j)\in F} \rho\left(\|r_l(d^{c_i}, X)\|^2_{\Sigma_l}\right) \tag{6}$$

$r_1$, $r_2$ denotes to the visual reprojection and IMU pre-integration residual factor; $r_3$, $r_4$ refers to the GICP relative pose and depth-to-map point-to-plane residual, respectively. $p^{c_i}$ represents the visual observation of the $i$-th frame; $z_{b_i b_{i+1}}$ denotes the pre-integrated observation; $I^{c_i}_{c_j}$ represents the GICP relative pose between the $i$-th and $j$-th frames; $d^{c_i}$ is defined as the depth-to-map observation of the $i$-th frame. $\rho$ is a core function to enhance the stability of the optimization step and mitigate the impact of outliers. In local BA, the $X$ signifies the IMU state $x$ of the keyframe and the visual map points $P$ while in single-frame optimization it only signifies the IMU state of adjacent frames.

#### 1) Extraction Plane Feature Points from Depth

Similar to SSL-SLAM [30], we extract plane feature points from point clouds to save computational resources, as direct depth-to-map registration can be computationally expensive due to the large volume of point cloud data generated by solid-state LiDAR. Initially, point clouds are segmented into a 2-D matrix by calculating the vertical angle $\alpha_i$ and horizontal angle $\theta_i$ of each point $p_i = (x_i, y_i, z_i,)$.

$$\alpha_i = \arctan\left(\frac{y_i}{x_i}\right) \tag{7}$$

$$\theta_i = \arctan\left(\frac{z_i}{x_i}\right) \tag{8}$$

Subsequently, these angles are used to equally divide the detection range, segmenting the point cloud into $M \times N$ cells. Within each cell, we calculate the geometric center of all points and define local smoothness to extract plane features. A lower smoothness value suggests a flat plane feature. This approach is more computationally efficient and suitable for handling the vast amount of point cloud data produced by solid-state LiDAR.

#### 2) Depth-to-map residuals

The depth residuals are generated by matching the sampled planar point cloud to the nearest submap. Keyframes that are within a specified spatiotemporal proximity to the current frame are identified within the local mapping thread. Subsequently, a submap is constructed utilizing the poses of these keyframes and their corresponding planar feature points.

Motivated by LIOSAM, The point-to-plane residual $r_l$ from the plane points to the local map is constructed as:

$$r_l(d^{c_i}, X) = (n^w)^T\left(P' - q^w_{plane}\right) = JX + \omega \quad (9)$$

$$J = \frac{\partial r_l}{\partial X} = \frac{\partial r_l}{\partial P'}\frac{\partial P'}{\partial X} = (n^w)^T[I, -P'^\wedge] \quad (10)$$

$$P' = T^w_c p^c_{plane} \quad (11)$$

where $\boldsymbol{n}^w$ denotes the normal vector of the plane in the world coordinate system, $q^w_{plane}$ represents the center point of the plane in the world coordinate system, $p^c_{plane}$ indicates the plane feature point in the camera coordinate system, and $T^w_c$ signifies the pose transformation matrix from the camera coordinate system to the world coordinate system. $\omega$ is the point-to-plane observation noise.

## V. Experiment

### A. Evaluation of GeoFlow-SLAM on Go2 Robot Dog and OpenLORIS Benchmark.

To evaluate our method in real-world scenarios, we conducted validation on the Go2-legged robot dog and G1 Humanoid Robot produced by Unitree, as depicted in Figure 5. . The robot dog is equipped with a D435i camera operating at about 10-15 FPS, Bosch BMI055 IMU in D435i at 200FPS , a XT16 LiDAR at 10 FPS, and a legged odometry system at 500 FPS. The G1 Humanoid Robot is equipped with a D435i camera operating at about 25-30 FPS, a Mid360 LiDAR at 10 FPS, and legged odometry at 500 FPS. The enhanced LIOSAM for the legged robot (similar to Leg-KILO) is employed to provide the ground truth. The trajectories from LIO-SAM and legged odometry are aligned to approximately estimate the extrinsic between the IMU and the legged odometry.

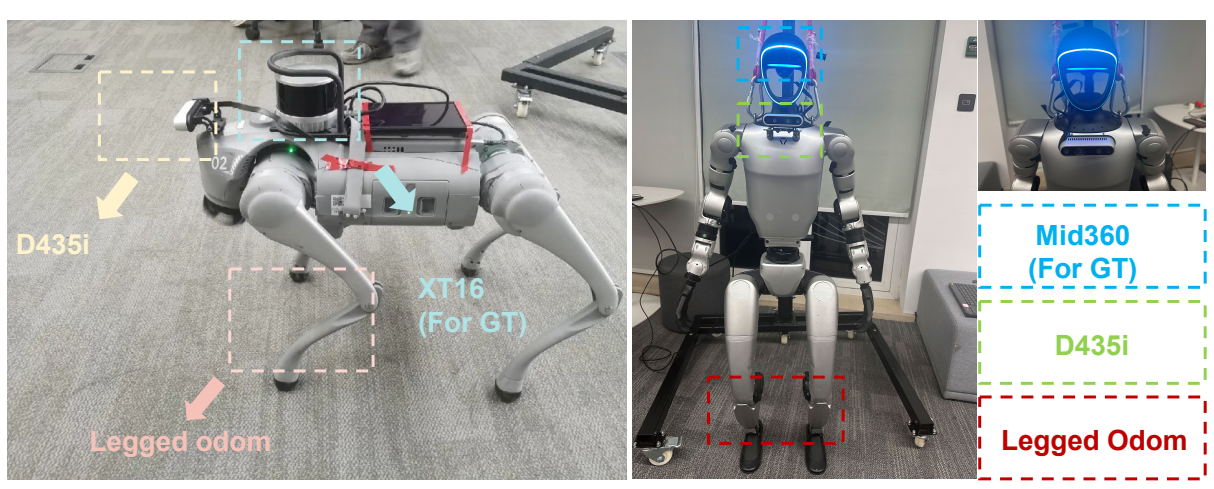

Figure 5. The Go2 legged robot dog (left) and the G1 Humanoid Robot (right).

We collected a total of six test sequences for the legged robot, four for robot dog and two for the humanoid robot as depicted in Figure 6. . In Seq1, the robot walked into an indoor office. The Seq2 recorded the robot exploring the very challenging texture-less corridor environment. Seq3 depicts a below-ground parking area, distinguished by its relatively low-light ambiance and reflection. To further validate the effectiveness of our algorithm in outdoor scenarios, we specifically acquired a dataset from an open plaza area, which presents challenges due to repetitive ground textures, as illustrated in Seq4. The Seq5 were recorded in an indoor office and Seq6 were in outdoor.

It is important to note that all sequences exhibited significant fast motion jitter during the robot's movement, resulting in image blurring and IMU distortion.

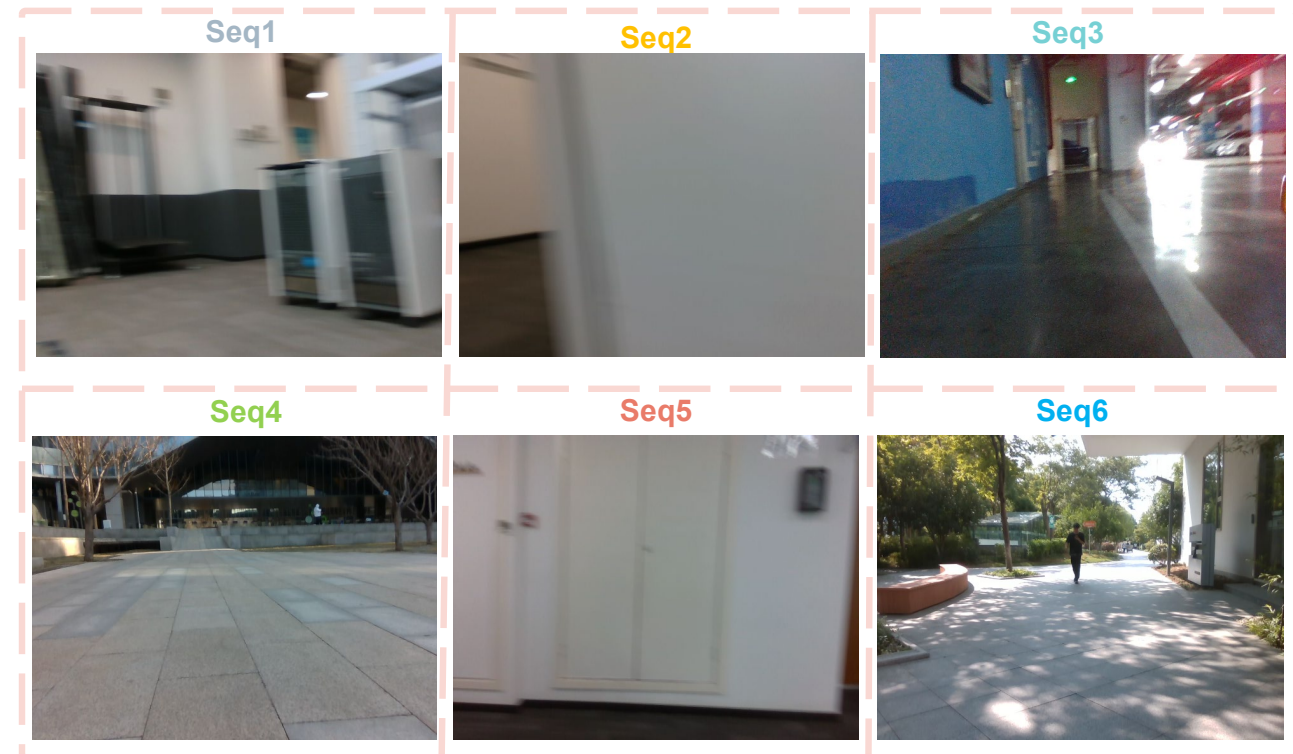

Figure 6. The Real-world challenge environments.Seq1-4 on Go2-legged robot dog and Seq5-6 on G1 humanoid robot

We compared our GeoFlow-SLAM with the VINS series (VINS-RGBD (RGBD+IMU) and S-VIO (RGBD+IMU)) and ORB-SLAM3, as they are among the most important works in the field of VSLAM. The Absolute Trajectory Error (ATE) results are listed in TABLE I.

TABLE I. ATE(M) EVALUATION RESULTS ON GO2 D435I DATASET.

| | VINS-RGBD | S-VIO | ORBSLAM3 | Ours |
|---|---|---|---|---|
| Seq1 | 0.8316 | 1.3621 | \ | **0.0947** |
| Seq2 | \ | \ | \ | **0.1369** |
| Seq3 | \ | \ | \ | **0.6009** |
| Seq4 | \ | \ | \ | **0.4458** |
| Seq5 | 0.3223 | 0.7121 | 0.2963 | **0.1709** |
| Seq6 | \ | \ | \ | **0.6235** |

TABLE II. RUNNING TIME(AVERAGE TIME PER-FRAME IN MS) ON 13TH GEN INTEL(R) CORE(TM) I9-13900H.

| | Seq1 | Seq2 | Seq3 | Seq4 | Seq5 | Seq6 |
|---|---|---|---|---|---|---|
| Ours | 24.9 | 26.5 | 27.6 | 26.3 | 20.1 | 27.1 |
| ORBSLAM3 | \ | \ | \ | \ | 9.0 | 11.6 |

In Seq1, our algorithm demonstrated significant improvements in positioning accuracy, achieving 88% enhancement over VINS-RGBD and 93% enhancement over S-VIO. Furthermore, across the Seq2, Seq3, Seq4 and Seq6 datasets, our algorithm was the only successful one, while other approaches encountered failures (always reset). In humanoid robot dataset Seq5, 45% improvement is achieved compared to ORBSLAM and our GeoFlow-SLAM has the best performance. As for running time, as shown in TABLE II. , our approach achieves real-time efficiency, processing each frame in approximately 20-30ms. However, due to the incorporation of additional observations, the computational efficiency is inevitably reduced compared to ORB-SLAM3. In Seq6, due to a tracking failure in the middle of the sequence by ORB-SLAM3, only the successfully tracked portion is used to compute the runtime.

Additionally, as shown in Figure 7. , our method achieves the highest consistency with the reference trajectory and the minimum end-to-end error. In contrast, the trajectories of VINS-RGBD and S-VIO are compromised by jumps and drift. Figure 8. illustrates the depth mapping results of the GeoFlow-SLAM on the two long indoor and outdoor datasets, Seq2 and Seq4.

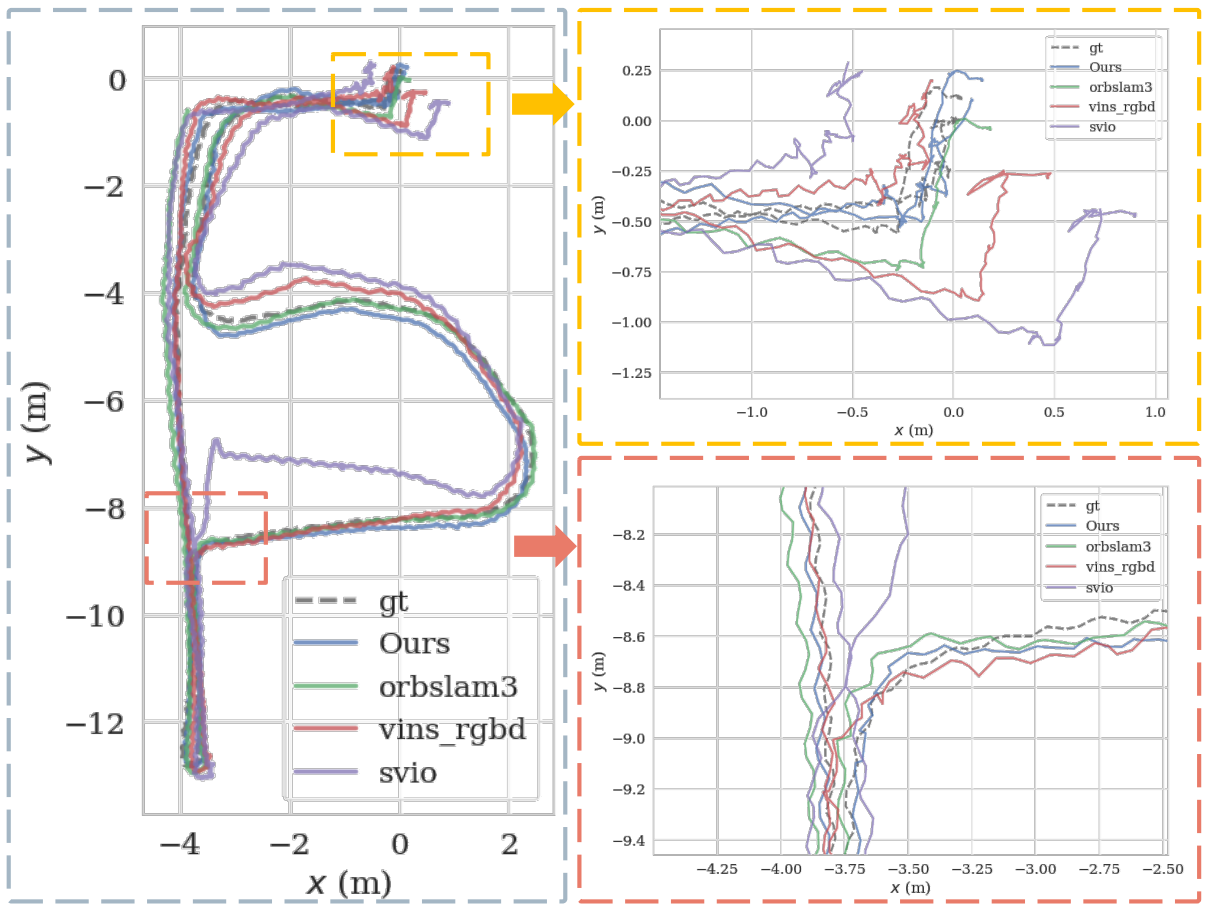

Figure 7. Estimated trajectory (colored) and ground truth (black) in the G1 Seq5 dataset.

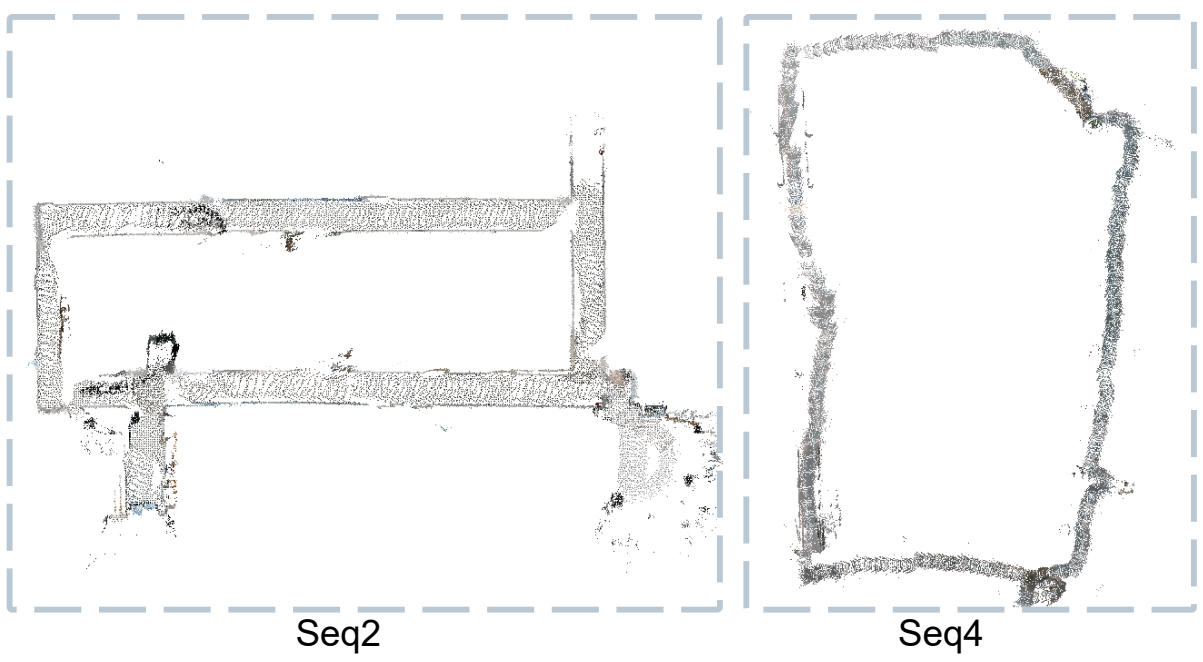

Figure 8. Our global plane points feature map in Seq2 and Seq4.

This superior performance can be attributed to three main factors:

**Robust Pose Initialization**: Due to the motion characteristics of legged robots, IMU data frequently suffers from errors. As shown in Figure 9. , the raw IMU data in Seq1-6 are severely distorted, with the maximum acceleration reaching up to 40 $m/s^2$ in robot dog and 20 $m/s^2$ in humanoid robot.

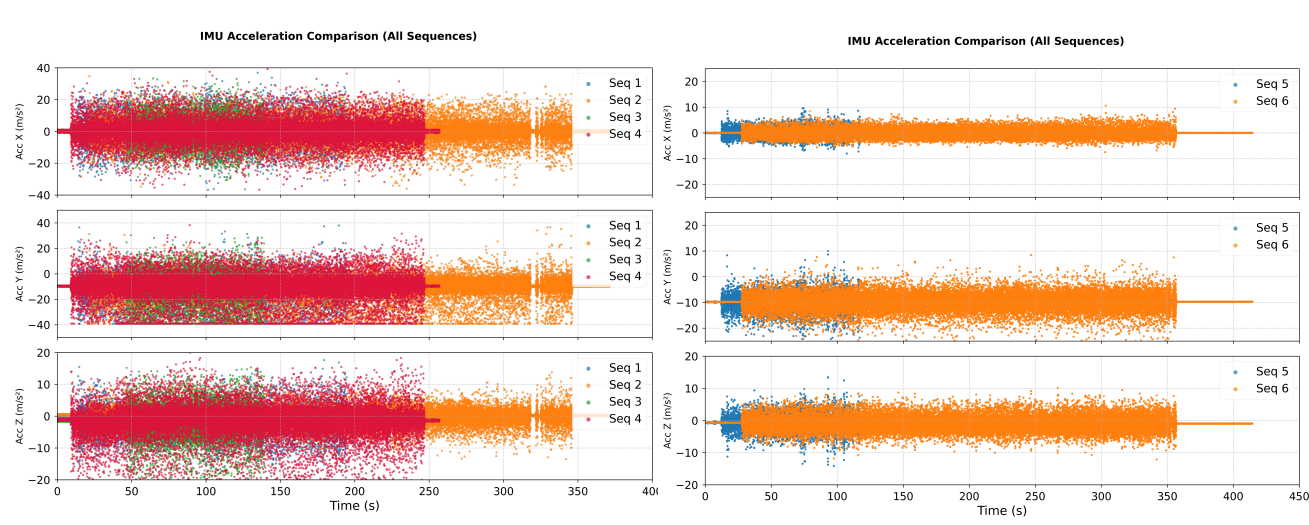

Figure 9. The raw IMU acceleration of our legged robot dog in Seq1-4 and humanoid robot in Seq5-6.

VINS-RGBD and SVIO utilize the IMU for pre-integration and pose prediction when visual and depth information are unavailable. These distortions can lead to optimization divergence. Unlike other methods that are heavily dependent on the IMU for initial pose estimation, accurate initialization of the robot's pose can still be achieved through the Legged, PnP, and GICP Odometry even in the absence of an IMU or when the IMU data distortions.

**Enhanced Dual-Stream Optical Flow Tracking**: The unique motion characteristics of legged robots often lead to image blur, which can severely impact the robustness of feature tracking. However, the dual-stream optical flow method we employ leverages prior map points and pose information, enabling effective feature tracking even when the current image is blurry. By projecting prior map points, we can continue tracking feature points effectively under such conditions. Additionally, increasing the number of visual feature points, particularly in low-texture environments, greatly improves the system's accuracy.

It is important to note that the Seq2 and Seq6 dataset includes dynamic objects, such as pedestrians and cars. Following optical flow tracking, our algorithm employs the fundamental matrix to filter out outliers that do not conform to the motion constraints. As illustrated in Figure 10. , this process mitigates the impact of dynamic objects on the algorithm to a certain extent.

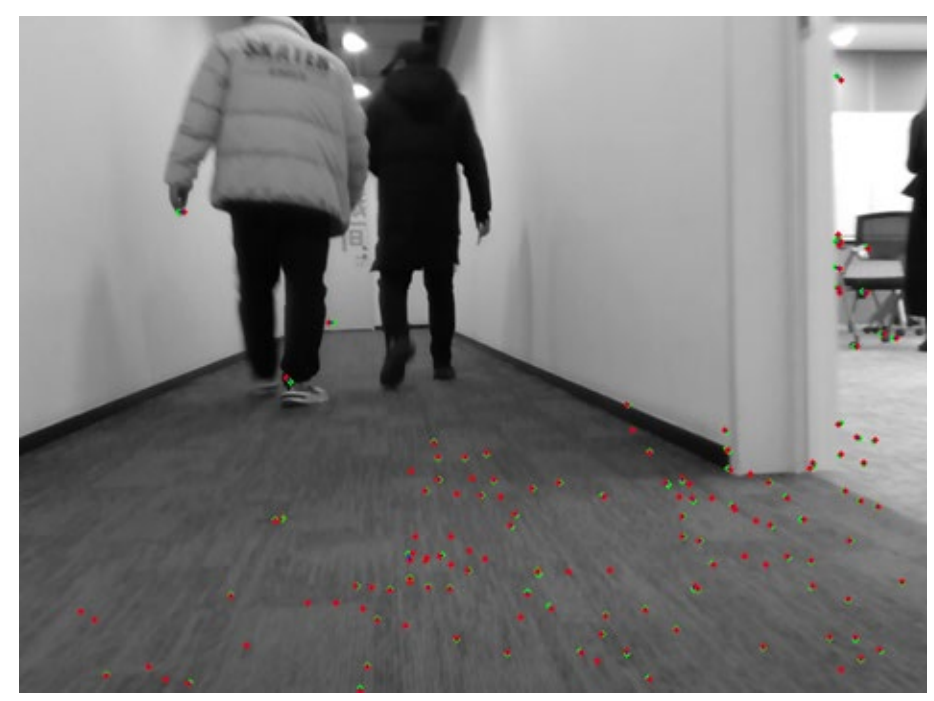

Figure 10. Feature points tracking result based on dual-stream optical flow in Seq2 with pedestrians. The green points denote feature points, while the red points indicate the reprojected points onto the image.

**Tightly Coupled Depth Constraints**: In texture-less environments, visual feature points are sparse, making pose estimation challenging. By coupling depth-to-map and GICP constraints, our method improves both accuracy and robustness. For example, in Seq2, a low-texture corridor, visual data is limited, but geometric features like the floor and walls provide valuable constraints, boosting robustness.

The experimental validations were conducted using the OpenLoris dataset, which was collected by a wheeled robot operating in various indoor environments, including residential and office spaces. These environments posed several challenges, such as low texture contrast, dynamic objects, and inadequate illumination. The results were systematically compared with several established VIO algorithms, including VINS-Mono (monocular + IMU), VINS-RGBD (RGB-D + IMU), ORB-SLAM3 (RGB-D + IMU), S-VIO (RGB-D + IMU), VID-SLAM (RGB-D + IMU), and DDIO-VIO (RGB-D + IMU). The ATE values of each VIO method are listed in TABLE III. The results indicate that by incorporating additional constraints such as optical flow and depth information, our method demonstrates enhanced robustness and accuracy across the majority of the OpenLoris datasets.

TABLE III. EVALUATION RESULT ON OPENLOROS DATASET. THE BEST RESULT IS HIGHLIGHTED IN BOLD AND SYMBOL \ INDICATES FAILURE.

| | VINS-Mono | VINS-RGBD | ORBSLAM3 | S-VIO | VID-SLAM | DDIO-VIO | GeoFlow-SLAM |
|---|---|---|---|---|---|---|---|
| Home1-1 | 1.0760 | 0.4520 | 0.9290 | 0.4020 | **0.3820** | 0.4060 | 0.4575 |
| Home1-2 | 0.5420 | 0.3790 | \ | **0.3160** | 0.3620 | 0.3470 | 0.3775 |
| Office1-1 | 0.2310 | 0.1010 | 0.0680 | 0.1010 | 0.0670 | 0.0780 | **0.0577** |
| Office1-2 | 0.2890 | 0.1240 | 0.1230 | 0.0990 | 0.1080 | 0.0700 | **0.0700** |
| Office1-3 | 0.1550 | 0.1540 | \ | 0.1510 | **0.0220** | 0.1040 | 0.0291 |
| Office1-4 | 0.3920 | 0.1860 | 0.2530 | 0.1640 | **0.1290** | 0.1440 | 0.1554 |
| Office1-5 | 0.2380 | 0.2390 | \ | 0.2120 | 0.2090 | 0.2140 | **0.2000** |

### B. *Ablation Study*

To evaluate the effectiveness of the proposed algorithm, ablation studies were conducted using nine sequences from the ScanNet dataset (50, 59, 84, 106, 169, 181, 207, 580, 616) and seven sequences from the TUM dataset (long_office_house, nostr_tex_far, nostr_tex_near, str_notex_far, str_notex_near, str_tex_far, str_tex_near), with various configurations of the algorithm implemented. A comprehensive assessment of performance across multiple test scenarios was carried out in RGBD mode without IMU and Legged odometry. The configurations employed in this study are categorized as follows: r+o represents the version utilizing only optical flow, r+i denotes the version employing only GICP, r+d indicates the version reliant solely on depth-to-map constraints, and r+o+i+d signifies the comprehensive version with all functionalities enabled. As illustrated in TABLE IV. , the incorporation of dual-stream optical flow, GICP, and depth-to-map constraints resulted in significant improvements in accuracy compared to ORB-SLAM3 in RGB-D mode.

On the ScanNet dataset, the mean ATE for the r+o+i+d configuration was 0.0779 meters, representing a 70% improvement in accuracy over ORB - SLAM3, which had an ATE of 0.2638 meters. Notably, ORB-SLAM3 failed to track in the sequences 50, 84, 181, 207, and 616. In contrast, the configurations of our algorithm (r+o, r+i, r+d, and r+o+i+d) did not experience any tracking failures across all sequences. Similarly, on the TUM dataset, all our configurations (r+o, r+i, r+d, and r+o+i+d) successfully tracked all sequences without failure, whereas ORB-SLAM3 failed to track in the texture-less and structure-less scenes, namely nostr_tex_far, str_notex_far, and str_notex_near. Moreover, all our configurations demonstrated varying degrees of improvement in positioning accuracy. Specifically, the r+o+i+d configuration achieved ATE and the Relative Trajectory Error (RTE) values of 0.0129 m and 0.0086 m, respectively, corresponding to improvements of 62.2% and 36.7% compared to ORB-SLAM3.

From the above results, it is evident that incorporating dual-stream optical flow, GICP initialization, and depth-to-map constraints effectively enhances the system's accuracy and robustness. The configuration that integrates all these constraints (r+o+i+d) performs the best.

TABLE IV. AVERAGE ATE AND RTE COMPARISON ON THE SCANNET AND TUM DATASETS.THE BEST RESULT IS HIGHLIGHTED IN BOLD

| | | orbslam3 | r+o | r+i | r+d | r+o+i+d |
|---|---|---|---|---|---|---|
| ScanNet | ATE | 0.2638 | 0.2404 | 0.1055 | **0.0775** | 0.0779 |
| | RTE | 0.0101 | 0.0095 | 0.0100 | 0.0094 | **0.0090** |
| TUM | ATE | 0.0341 | 0.0142 | 0.0156 | 0.0155 | **0.0129** |
| | RTE | 0.0136 | 0.0089 | 0.0106 | 0.0100 | **0.0086** |

A comparative analysis was further conducted between our algorithm and the SOTA RGBD-SLAM method, STL-SLAM[18]. The results in TABLE V. indicate that our approach attains enhanced accuracy.

TABLE V. STL-SLAM, ORBSLAM3 COMPARISION OF ATE (M) ON TUM. {\} INDICATE FAILURE

| | STL-SLAM | ORBSLAM3 | GeoFlow-SLAM |
|---|---|---|---|
| fr1/xyz | **0.0100** | 0.0105 | 0.0101 |
| fr1/desk | 0.0280 | 0.0164 | **0.0162** |
| fr3/cabinet | **0.0190** | \ | 0.0500 |
| fr3/large_cabinet | 0.0620 | \ | **0.0601** |
| fr3/str_notex_far | 0.0330 | \ | **0.0116** |
| fr3/str_notex_near | 0.0130 | \ | **0.0126** |
| fr3/str_tex_far | 0.0140 | 0.0102 | **0.0089** |
| fr3/str_tex_near | 0.0110 | **0.0093** | 0.0094 |

### C. *Extremely Hard Scene Experiments*

Due to the integration of depth-to-map constraints and a dual-stream optical flow tracking mechanism, our algorithm demonstrates robustness in environments characterized by low texture and low structural complexity. To further validate the effectiveness of our approach, we conducted additional tests on two particularly challenging scenarios within the no_structure_notexture_far and nostructure_notexture_near in the TUM dataset. These scenarios are known for their lack of texture and structure. The results show our algorithm is highly resilient and accurate in texture-less environments.

In situations where texture is sparse and patterns are repetitive, traditional optical flow tracking methods often encounter difficulties. However, our algorithm utilizes precise pose priors to mitigate incorrect feature point matching. Even in the absence of visual observations, the depth-to-map point-to-plane constraints and the GICP relative pose constraints serve to effectively govern the pose estimation, thus preventing the optimization process from diverging. The ATE metrics for the nostr_notex_far and nostr_tex_near datasets are 0.0308 meters and 0.0467 meters, respectively. As shown in Figure 11. , in nostr_notex_far, our SLAM still constructs the global plane feature map.

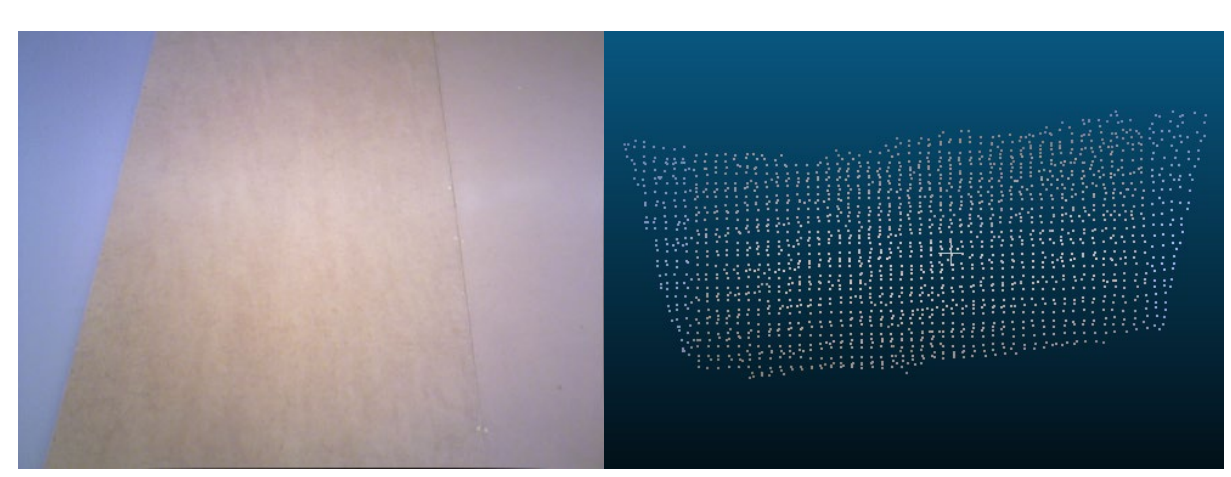

Figure 11. Our global plane feature map in Nostr_notex_far.The left is the original image and the right is our global plane feature point cloud.

## VI. CONCLUSION

In this study, we propose GeoFlow-SLAM, a novel SLAM system for legged robots, achieving significant improvements in feature point matching, pose initialization, and overall system robustness and accuracy. Specifically, we have introduced a method that integrates dual-stream optical flow tracking with prior map points and pose information, enhancing feature tracking in fast locomotion scenarios. Additionally, we have developed a pose initialization method combining IMU/Legged Odometry, inter-frame PnP, and GICP, ensuring reliable pose estimation even without IMU data. Furthermore, tightly coupling the geometric GICP constraints, our method significantly improves performance in visually texture-less environments. GeoFlow-SLAM achieves sota performance on multiple benchmark datasets. To facilitate further research, we will release the code and legged robotics datasets.

Moving beyond geometric SLAM, our future work will focus on scene graph-based semantic reconstruction to provide structured, object-centric representations for legged robotics.